\documentclass[lettersize,journal]{IEEEtran}
\usepackage{amsmath,amsfonts}
\usepackage{algorithmic}
\usepackage{array}
\usepackage[caption=false,font=normalsize,labelfont=sf,textfont=sf]{subfig}
\usepackage{textcomp}
\usepackage{stfloats}
\usepackage{url}
\usepackage{verbatim}
\usepackage{graphicx}
\def\BibTeX{{\rm B\kern-.05em{\sc i\kern-.025em b}\kern-.08em
    T\kern-.1667em\lower.7ex\hbox{E}\kern-.125emX}}
\usepackage{balance}
\begin{document}
\title{BdSLW401: Transformer-Based Word-Level Bangla Sign Language Recognition Using Relative Quantization Encoding (RQE)}
\author{Husne Ara Rubaiyeat, Njayou Youssouf, Md Kamrul Hasan, Hasan Mahmud
\thanks{}}


\maketitle

\begin{abstract}
Sign language recognition (SLR) for low-resource languages like Bangla suffers from signer variability, viewpoint variations, and limited annotated datasets. In this paper, we present BdSLW401, a large-scale, multi-view, word-level Bangla Sign Language (BdSL) dataset with 401 signs and 102,176 video samples from 18 signers in front and lateral views. To improve transformer-based SLR, we introduce Relative Quantization Encoding (RQE), a structured embedding approach anchoring landmarks to physiological reference points and quantize motion trajectories. RQE improves attention allocation by decreasing spatial variability, resulting in 44.3\% WER reduction in WLASL100, 21.0\% in SignBD-200, and significant gains in BdSLW60 and SignBD-90. However, fixed quantization becomes insufficient on large-scale datasets (e.g., WLASL2000), indicating the need for adaptive encoding strategies. Further, RQE-SF, an extended variant that stabilizes shoulder landmarks, achieves improvements in pose consistency at the cost of small trade-offs in lateral view recognition. The attention graphs prove that RQE improves model interpretability by focusing on the major articulatory features (fingers, wrists) and the more distinctive frames instead of global pose changes. Introducing BdSLW401 and demonstrating the effectiveness of RQE-enhanced structured embeddings, this work advances transformer-based SLR for low-resource languages and sets a benchmark for future research in this area.

\end{abstract}


\section{Introduction}
\IEEEPARstart{T}{he} study of sign language recognition (SLR) is vital to the communication and interaction with the deaf and hearing communities. Recently SLR has improved considerably due to deep learned techniques, transformer-based models to be particular. The self-attention technique \cite{Vaswani2017} has proven to be the solution of many issues in natural language processing (NLP) as well as vision-related tasks \cite{Bahdanau2015}. However, where to apply attention is just as important as how it is applied. Often, when raw unstructured landmark data is fed to transformers, each part of the model is allocated attention inappropriately. This is ineffective for learning features. Similar to people who have trouble with understanding high-density information, models can take advantage of structured input representations that target salient sign gestural patterns while getting rid of redundancy.

Datasets contribute to robust recognition models like BdSL which remains unrepresented in large-scale datasets. Other datasets like Word-Level American Sign Language (WLASL) \cite{li2020word}, a new large-scale multi-modal Turkish Sign Language dataset (AUTSL) Turkish \cite{Sincan2020}, and SignBD \cite{Rahman2023} helped advance transformer based SLR, but BdSL research has used smaller datasets with feature extraction \cite{Rubaiyeat2024}.

These methods struggle with signer variability, changes in viewpoint, and variable gesture forms, limiting their viability for real-world deployment. To fill this gap, we present BdSLW401, a novel large-scale word-level multiview Bangla Sign Language (BdSL) dataset that contains 401 unique signs, 102176 video samples spanning across 18 different signers, front and lateral view, gesture boundary annotations, and a diverse group of signer morphology. We provide a benchmark for transformer-based SLR models to validate the datasets, as it goes beyond the traditional handcrafted feature-based approaches.

One key limitation in current SLR systems is that raw landmark embeddings introduce unnecessary variations due to hand dominance, body proportions, and camera angles. Transformers that process these raw embeddings struggle to allocate attention effectively. Inspired by human cognitive strategies, where perception organizes sensory input before attention is applied, we propose Relative Quantization Encoding (RQE).

RQE enhances transformer attention by anchoring to physiological reference points (e.g. wrist for fingers, shoulders for torso) and quantizing the MediaPipe-provided landmarks. It enables a more effective attention distribution, allowing transformers to focus on meaningful motion patterns rather than redundant noise. Additionally, we introduce RQE-SF (Shoulder-Freezing), which stabilizes upper body reference points, ensuring pose-consistent embeddings across different signers and viewpoints.

We benchmark BdSLW401 using SLRT (Sign Language Recognition Transformer) \cite{Bohacek2022} and evaluate its impact on BdSLW60, WLASL, and SignBD. Our results confirm that RQE significantly improves recognition accuracy, reduces the word error rate (WER) by 1.2–44.3$\%$, achieving 84.48$\%$ accuracy on BdSLW60, and improving attention distribution. Attention heat maps demonstrate that RQE-trained models focus more effectively on finger articulations and distinctive frames, while models using raw landmarks distribute attention inconsistently. However, performance gains diminish in ultra-large datasets (e.g., WLASL2000), suggesting the need for adaptive noise handling strategies in future research.

The main contributions of this paper are: 

\begin{itemize}
    \item BdSLW401: A Large-Scale Bangla Sign Language Dataset, featuring 401 signs, multi-view recordings, and manual gesture annotations.
    \item RQE: A Structured Input Encoding Method, improving transformer attention by reducing noise and enhancing feature extraction.
    \item Cross-Dataset Validation, demonstrating RQE enhances recognition on BdSLW60, WLASL, and SignBD.
\end{itemize}

The remainder of this paper is structured as follows: Section \ref{RelatedWork} reviews related work on SLR datasets, transformer-based recognition models, and landmark encoding methods. Section \ref{BdSLW401Dataset} introduces BdSLW401, detailing its collection process, annotation, and comparison with existing datasets. Section \ref{Method} presents RQE and its integration into transformers. Section \ref{Experiments} discusses experiments, evaluation metrics, and benchmarking results. Section \ref{Conclusion} analyzes RQE’s impact on attention mechanisms, real-time feasibility, and generalization. Section 7 concludes the paper with key findings and future directions.

\section{Related Work}
\label{RelatedWork}
Large-scale datasets, deep learning architectures, and structured feature representations have driven recent advances in sign language recognition (SLR). While transformer-based models have improved recognition accuracy, they struggle with raw pose-based input due to signer variability, viewpoint shifts, and unstructured landmark embeddings. Below, we discuss existing SLR datasets, recognition techniques, and structured input representations, highlighting their limitations. These limitations serves as the foundation for introducing our BdSLW401 dataset and the RQE-based approach.

\subsection{Word-Level Sign Language (WLSL) Datasets}
The progress in sign language recognition (SLR) largely depends on large-scale, well-annotated datasets. Among the most influential datasets, WLASL \cite{li2020word} has significantly contributed to word-level American Sign Language (ASL) recognition, yet it lacks viewpoint diversity. 

AUTSL \cite{Sincan2020} focuses on Turkish Sign Language (TSL) and introduces multimodal RGB-D input, making it more adaptable to deep learning models. Similarly, INCLUDE \cite{sridhar2020include} is an Indian Sign Language (ISL) dataset that addresses class imbalance issues, but high inter-class similarity makes generalization difficult. For Bangla Sign Language (BdSL), the available datasets are still limited in scale and variability. SignBD \cite{Rahman2023} was one of the first BdSL datasets based on OpenPose features using deep learning models. BdSLW60 \cite{Rubaiyeat2024} improved upon this by introducing pose-based embeddings, yet it remains constrained in vocabulary size (60 signs) and lacks multi-view recordings. These constraints prevent BdSL models from achieving high generalization across diverse signer representations. A comparison of existing word-level sign language dataset is gvein in Table~\ref{tab:dataset_summary}.
\begin{table*}[t]
\centering
\caption{Comparison of Word-Level Sign Language Datasets}
\label{tab:dataset_summary}
\begin{tabular}{|l|l|l|l|c|c|c|c|}
\hline
\textbf{Dataset} & \textbf{Language} & \textbf{Modality} & \textbf{Baseline Model} & \textbf{\# Signs} & \textbf{\# Videos} & \textbf{\# Signers} & \textbf{Year} \\ \hline
WLASL \cite{li2020word}  & ASL    & RGB Video & I3D, CNN-LSTM & 100-2000 & ~21,083 & ~119 & 2020 \\ \hline
AUTSL \cite{Sincan2020}  & Turkish & RGB-D (Kinect) & CNN-LSTM, 3D ResNet & 226 & 38,336 & 43 & 2020 \\ \hline
SignBD \cite{Rahman2023}  & Bangla  & RGB Video & CNN, LSTM & 200 & 6,000 & 16 & 2023 \\ \hline
BdSLW60 \cite{Rubaiyeat2024}  & Bangla  & Landmark (MediaPipe) & Attention-based Bi-LSTM & 60 & 9,307 & 18 & 2024 \\ \hline
INCLUDE \cite{sridhar2020include} & Indian  & RGB Video & CNN, RNN & 263 & 4287 & 7 & 2020 \\ \hline
\textbf{BdSLW401 (Ours)} & Bangla  & Landmark (MediaPipe) & SLRT + RQE & 401 & 102176 (Front: 51098, Lateral: 51078) & 18 & 2024 \\ \hline
\end{tabular}
\end{table*}

To address these issues, we introduce BdSLW401, the first large-scale BdSL dataset with 401 word signs, 102,176  samples from 18 signers, covering both frontal and lateral viewpoints. BdSLW401 provides gesture boundary annotations, pose-based embeddings, and a benchmark for transformer-based recognition models, filling the gap in BdSL research.


\subsection{Transformer-Based Sign Language Recognition}

With the introduction of self-attention mechanisms, transformers have shown remarkable success in sequential learning \cite{Vaswani2017}. In SLR, Camgoz et al. \cite{camgoz2020sign} pioneered transformer-based SLR, yet their performance dropped when using raw, unprocessed landmarks due to noise and inconsistencies. SignPose transformer \cite{Bohacek2022} further demonstrated that transformers outperform CNN-RNN pipelines in word-level recognition. However, a key limitation remains—raw pose-based embeddings introduce unwanted variations from signer body proportions, hand dominance, and camera angles, leading to suboptimal attention allocation.

Despite advances in pose-based sign recognition, existing models often overlook the importance of structured feature representation. While MediaPipe Holistic \cite{lugaresi2019mediapipe} and OpenPose \cite{cao2017realtime} enable efficient pose extraction, they inherit depth distortions and signer-specific biases, which can mislead transformer models. Although Bohacek et al. \cite{Bohacek2022} investigated the application of pose-based transformers to tackle this issue, their approach lacked systematic normalization procedures. When applied to different signers and camera angles, this absence led to differences in embedding consistency, exposing serious problems with regard to generality in their framework.

\subsection{Landmark-based representation and Quantization}

Keypoint detection frameworks, like MediaPipe Holistic \cite{lugaresi2019mediapipe} and OpenPose \cite{cao2017realtime}, are primarily used in modern sign language recognition systems. These frameworks capture anatomical landmarks across hands, facial features, and torso regions to reduce processing requirements and improve motion modeling capabilities. While this method avoids the noise present in raw video streams, it introduces basic differences due to unique signing styles, different camera angles, and artifacts in spatial perception \cite{Mahmud2024}. In transformer architectures, where self-attention layers commonly confuse semantically relevant movements with biomechanical variety and environmental noise, this phenomenon leads to embedding space distortions \cite{Ferreira2021}. Current mitigation techniques show some degree of effectiveness, such as cross-signer adaptation by meta-learning \cite{Ferreira2021} and spatial regularization through depth correction \cite{Mahmud2024}. The kinematic links between body segments are usually ignored by spatial regularization techniques, whereas meta-learned solutions are unable to create strong feature hierarchies, which severely restricts their ability to generalize for unseen signers.

Existing approaches' shortcomings are exacerbated by their disdain for human perceptual principles. Understudied in SLR, cognitive theories like Gestalt grouping principles \cite{Wertheimer1923} and Norman's Action Cycle \cite{Norman1988} highlight how structured input improves recognition efficiency and lowers cognitive strain. Similar to expecting people to understand gestures without context, raw pose embeddings process landmarks as absolute coordinates, requiring transformers to deduce spatial relationships from the ground up. Humans naturally sense motion by focusing motions to physiological reference points (e.g., wrists relative to shoulders) and filtering out noise (e.g., ignoring lower-body fluctuations during upper-body indications) \cite{Wertheimer1923}.

In order to combat this, we introduce Relative Quantization Encoding (RQE). RQE is a technique that combines technical pose normalization with a cognition-driven approach. In reference to dominant landmarks to stable physiological markers (e.g., torso or spine) and segmenting continuous motion trajectories into discrete interpretable levels, RQE is able to effectively eliminate signer-specific variation. 

The recognizer removes irrelevant information (e.g., lower-body movements) and merges semantically related keypoints (e.g., fingers with wrists) so that RQE has consistent spatial scaling over any signers and observations. \cite{Norman1988, Wertheimer1923}, and the strategies used by humans to perceive.

This method of systematic way of doing things minimizes "cognitive load" of transformers and also maximizes the efficiency of transformers in response to gesture-specific patterns. In contrast to typical “black-box” deep learning approaches that operate on unstructured landmarks, RQE enhances interpretability by mapping learned features onto physiologically pertinent categories. In contrast to previous work, RQE offers a direct mapping of transformer attention mechanisms to established cognitive principles. This enables models to attend to motion patterns, reducing spatial noise, when classifying visual stimuli.

RQE builds upon the principles of effective computing and the fundamentals of human perception. It establishes itself as a significant step towards generalized and scalable sign language recognition (SLR) systems.

\section{BdSLW401 Dataset}
\label{BdSLW401Dataset}
\subsection{Dataset Overview}
The BdSLW401 dataset is a leap forward for Bangla Sign Language recognition. It resolves the issues (e.g lack of viewpoint diversity, dataset scale) in earlier resources like BdSLW60. This dataset is captured from front and lateral views and prioritizes natural movement patterns by capturing 401 distinct BdSL gestures from 18 signers. Instead of tracking facial expressions, it tracks 33 body posture features and 21 accurate hand landmarks per hand. This leads to a more concrete focus on gesture dynamics.

As opposed to more conventional methods that are time-consuming frame-by-frame editing, we apply a "whole-cut" approach. Video clips are trimmed to start when hand(s) initially move and stop when they leave the frame, discarding unneeded information. To address inconsistencies caused by signer height, camera angles, or lighting, the dataset presents Relative Quantization Encoding (RQE). The method standardizes gestures under different recording conditions so that the system can handle real-world variation. With its emphasis on accuracy and scalability, BdSLW401 sets a new benchmark for sign language technology, a one that better replicates the way humans naturally communicate, without technical constraints.

Consequently, BdSLW401 provides a scalable and dependable basis for improving BdSL identification and acts as a strong benchmark for transformer-based models in low-resource environments.

\subsection{Data Collection Process}
\label{DataCollection}
To record the variations, 18 native BdSL signers—3 female, 15 male; ages 18–25—were selected from linguistically different areas of Bangladesh. Each signer performed 10–20 trials per word in unconstrained setting. The detail of the data collection process is discussed in our previous paper \cite{Rubaiyeat2024}. Signers provided front and lateral view trials in a single sitting  in sequential manner.

Landmarks were extracted using MediaPipe Holistic \cite{lugaresi2019mediapipe}, focusing exclusively on 33 pose landmarks (11 face landmakrs, 10 lower body landmarks, shoulders, elbows, wrists, and a few hand landmarks) and 21 landmarks per hand, with 468 facial data excluded to prioritize manual articulators. To ensure consistency, local coordinate systems were anchored to physiological reference points: wrists for hand landmarks, mid-shoulder (first frame) for upper-body points, and the same approach was followed for the lower-body landmarks.

\subsection {Annotation Protocol and Standardization}
Annotation followed a three-stage process to ensure quality and consistency. 

First, manual cropping segmented clips using a multi-marker tool, adopting a "whole-cut" strategy to handle challenges like unsynchronized motions, unintentional movements, and signs requiring wrist withdrawal. This approach diverged from BdSLW60's precise frame-level cropping but improved scalability. Two annotators cross-verified all clips in fast-forward mode, discarding around 15\% of samples due to occlusions or incomplete signs. 

Second, sample restriction capped the number of retained samples to $\leq 10$ per signer-word pair, mitigating overrepresentation and ensuring statistical validity. 

Third, physiological coordinate normalization standardized landmarks by translating the global coordinates to physiological local coordinates as described in \ref{DataCollection}. Missing landmarks (e.g., occluded wrists) were set to zero to avoid quantization outliers.

\subsection{Dataset Statistics}
BdSLW401 is characterized by its scale, diversity, and structured encoding. The dataset includes 401 words spanning everyday communication, education, and healthcare, with deliberate inclusion of morphologically challenging groups such as near-identical pairs (e.g., W043 vs. W047, differing only in finger selection) and high-variability signs (e.g., W077’s two-handed asymmetry). Eighteen signers contributed 51,098 front-view and 51,078 lateral-view clips, with frame counts ranging from 7–370 (front) and 16–332 (lateral). 

Each frame is encoded as a 224 (75 × 3-1) tensor, comprising 33 pose landmarks minus last depth point and 21 landmarks per hand. Each landmark is represented as (x, y, d), representing the normalized x-coordinate, normalized y-coordinate, and normalized depth value, respectively.

\subsection{Comparison with Existing Datasets}
BdSLW401 addresses critical gaps in Bangla SLR research. Compared to BdSLW60, it offers 6.7× larger vocabulary and multi-view support, while avoiding annotation noise through sample restrictions. Unlike INCLUDE (263 words), which lacks structured encoding, BdSLW401 employs RQE to decouple gestures from signer-specific variability. It surpasses SignDB in diversity, with sign variations and challenging word pairs absent in smaller datasets. While WLASL (2,000 ASL words) and AUTSL (226 Turkish signs) include facial landmarks, BdSLW401’s motion-centric design prioritizes manual articulators, aligning with BdSL’s linguistic structure. 

\begin{table*}[t]
\centering
\caption{Comparison of BdSLW401 with Existing Datasets}
\label{tab:dataset_comparison}
\begin{tabular}{|l|c|c|c|c|c|}
\hline
\textbf{Feature}              & \textbf{BdSLW401} & \textbf{WLASL} & \textbf{BdSLW60} & \textbf{INCLUDE} & \textbf{SignBD} \\ \hline
\textbf{Language}             & Bangla            & ASL            & Bangla           & Indian           & Bangla                   \\ \hline
\textbf{Words}                & 401               & 2,000          & 60               & 263            & 200                      \\ \hline
\textbf{Views}                & Front + Lateral   & Front          & Front            & Front            & Front                    \\ \hline
\textbf{Signers}              & 18                & 119            & 18                & 7               & 16                       \\ \hline
\textbf{Landmarks}            & Pose + Hands      & Video           & Video             & Video             & Raw (Openpose)                      \\ \hline
\textbf{Quantization}         & RQE-based         & None           & RQ             & None             & None                     \\ \hline
\textbf{Cropping}             & Manual (whole-cut) & Manual           & Precise          & Manual           & Manual                   \\ \hline
\end{tabular}
\end{table*}

\subsection{BdSLW401 Dataset Splits for Training, Validation, and Testing}
Splits were designed to evaluate cross-signer generalization and view robustness. The test set includes fixed signers (S04, S08) across both views, ensuring evaluation on unseen users. A stratified 12\% validation split (1 trial per signer-word pair) was drawn from the remaining signers, with the rest allocated to training. This resulted in 38,876 front-view and 38,828 lateral-view training clips, 4,389 front and 4,380 lateral validation clips, and 7,833 front and 7,870 lateral test clips.

\begin{table}[h!]
\centering
\caption{Dataset Splits for BdSLW401}
\label{tab:dataset_splits}
\begin{tabular}{|l|c|c|c|}
\hline
\textbf{View}         & \textbf{Train} & \textbf{Validation} & \textbf{Test} \\ \hline
\textbf{Front}        & 38,876         & 4,389               & 7,833         \\ \hline
\textbf{Lateral}      & 38,828         & 4,380               & 7,870         \\ \hline
\end{tabular}
\end{table}

\subsection{Dataset Difficulty}
The BdSLW401 dataset is designed to rigorously evaluate model robustness by incorporating morphologically challenging word groups and signer-dependent variations inherent to real-world Bangla Sign Language (BdSL). These challenges are categorized as follows:

\subsubsection{Near-Identical Pairs with Sequential Variations}
A core challenge involves distinguishing words that share identical base gestures but differ in motion prefixes, postfixes, or directional reversals. For example, the group (W025, W030, W033, W034) uses a common circular wrist motion but varies in starting/ending positions or movement sequences. The group (W143, W144, W145, W147, W150, W154) contains similar characteristics. Pairs like W028 and W029 differ in the reversal of motion sequences and finger articulations. W027 is a submotion of W026, retaining only the last portion of the gesture. These subtle sequential variations test a model’s ability to parse temporal dependencies and kinematic differences.

\subsubsection{Minor Differences and use of Fingers}
Several groups require fine-grained discrimination of finger-level variations and also their position relative to the face. The pair W043 and W047, for instance, share identical motion trajectories but differ in the selected fingers. Words like W095, W375, and W380 further challenge models with the use of fingers around the face and hand positions. 

\subsubsection{Signer-Dependent Variability}
Regional dialects and individual signing styles introduce variability in how the same word is performed. For instance, W019 exhibits differences in hand arc height (high vs. low) across signers, while W077 varies in two-handed coordination (symmetrical vs. asymmetrical). W164 demonstrates mixed hand dominance, with some signers emphasizing the left hand and others the right. Compound motions like those in W311 (finger flicks combined with elbow rotations) also vary in speed and emphasis, reflecting signer-specific articulation.

\subsubsection{Left-Handed Samples and Mirroring Effects}
Approximately 17\% of the dataset features left-handed signing, introducing mirror-image variations that models must normalize (statistics from BdSLW60 \cite{Rubaiyeat2024}). 

\subsubsection{Confounding Factors}
Unintentional movements, such as head tilts in W102 (S05) can be aligned to W164, adding noise to motion patterns. Temporal ambiguities arise in signs like W034 and W070, where pauses or repeated motions complicate frame-level alignment.

\subsection{Input Embedding}
Landmarks are structured as 224 (75 × 3-1) tensor (x, y, d), representing the normalized x-coordinate, normalized y-coordinate, and normalized depth value, respectively. It comprises 33 pose points (excluding the last depth point), 21 left-hand landmarks, and 21 right-hand landmarks. The RQE framework encodes relative motion trajectories: hand landmarks are quantized relative to wrist positions, upper-body points to the mid-shoulder origin, and lower-body points to a fixed level. Initially we quantized the lower body parts accordingly to RQE and then fixed them to one single level, because, in sign language, the lower body movements are irrelevant. This structured approach reduces spatial redundancy, enabling transformers to focus on gesture-relevant patterns while maintaining compatibility with real-time inference requirements. 224 tensor size was chosen as the number is even and can be factorized to different multihead experiments.  

\section{Methodology}
\label{Method}
\subsection{Overview of Sign Language Recognition Transformer (SLRT)}
The Sign Language Recognition Transformer (SLRT) builds on the architecture proposed by Camgoz et al. \cite{camgoz2020sign}, which introduced joint sign language recognition and translation using self-attention mechanisms. Transformers leverage multi-head attention to capture long-range dependencies, making them well-suited for modeling complex temporal sequences in sign language recognition (SLR) \cite{Vaswani2017}. However, the original SLRT model relied on CNN-extracted features from raw video frames, which introduced computational bottlenecks due to the sequential nature of CNNs. According to SignPose transformer, CNN-based I3D can be 3-4 times computationally complex yet 10 times slower in reasoning \cite{Bohacek2022}. 

Our implementation eliminates this inefficiency by directly operating on landmark sequences, removing the need for CNN-based feature extraction. The model consists of a transformer encoder with three layers and eight attention heads, designed for parallel processing of variable-length sign sequences. Instead of processing raw spatial coordinates, inputs are transformed into 1D embeddings (224 dimensions) using sinusoidal positional encoding \cite{Vaswani2017}, ensuring temporal coherence across frames.

\begin{figure}[htbp]
    \centering
    \includegraphics[width=0.5\textwidth]{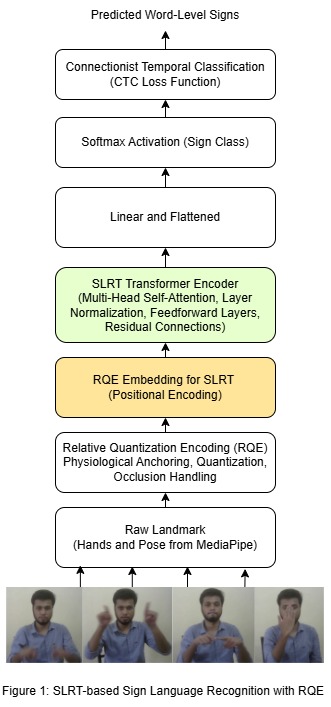} 
    \caption{SLRT-based Sign Language recognition with RQE.}
    \label{fig:SLRT_RQE_pipeline}
\end{figure}

By replacing CNN-based preprocessing with landmark-based inputs, our approach significantly improves computational efficiency to near real time while maintaining compatibility with the transformer’s self-attention mechanism \cite{camgoz2020sign}. However, raw landmark-based embeddings still introduce variability due to signer differences, depth distortions, and viewpoint inconsistencies \cite{Bohacek2022}. To mitigate these issues, Relative Quantization Encoding (RQE) is introduced to structure pose embeddings before feeding them into the transformer. The overall pipeline of our SLRT with RQE is shown in Figure \ref{fig:SLRT_RQE_pipeline}.

\subsection{Why standardization of hand and body landmarks: Theoretical perspective}
Transformer-based sign language recognition (SLR) requires consistent pose embeddings to ensure stable attention allocation. However, raw landmark sequences capture signer-specific proportions, leading to embedding inconsistencies across different users and viewpoints. To mitigate this, we implement landmark standardization, inspired by gesture normalization principles \cite{molchanov2016online}, to decouple gestures from individual signer variations.

Standardization follows three key transformations:

\begin{itemize}
    \item Hand Landmark Standardization → Hand coordinates are expressed as offsets from the wrist, ensuring consistent representation of finger positions across signers. This prevents arm length variations from affecting gesture interpretation (Figure \ref{fig:RQE_frames_level}).
    \item Upper-Body Normalization → To maintain relative arm positioning, upper-body landmarks (shoulders, elbows) are scaled using the first-frame shoulder width and torso length. This stabilizes arm motion trajectories, making embeddings less dependent on signer physique or posture shifts.
    \item Lower-Body Anchoring → While lower-body movements are rarely critical for manual gestures \cite{bragg2019sign}, their variability introduces unnecessary environmental noise. To suppress this, hip landmarks are anchored to the shoulders' midpoint, ensuring consistent global pose alignment.
\end{itemize}

For example, in gesture W380, 'hands up' motion is consistently represented across signers by expressing finger positions relative to the wrist, independent of arm length or posture variations. This aligns with human perceptual grouping principles \cite{Wertheimer1923}, ensuring that transformers focus on salient motion patterns rather than individual body variations.

\subsection{Relative Quantization Encoding (RQE) for Landmark-Based Representation}

\begin{figure*}[t]
    \centering
    \includegraphics[width=0.8\textwidth]{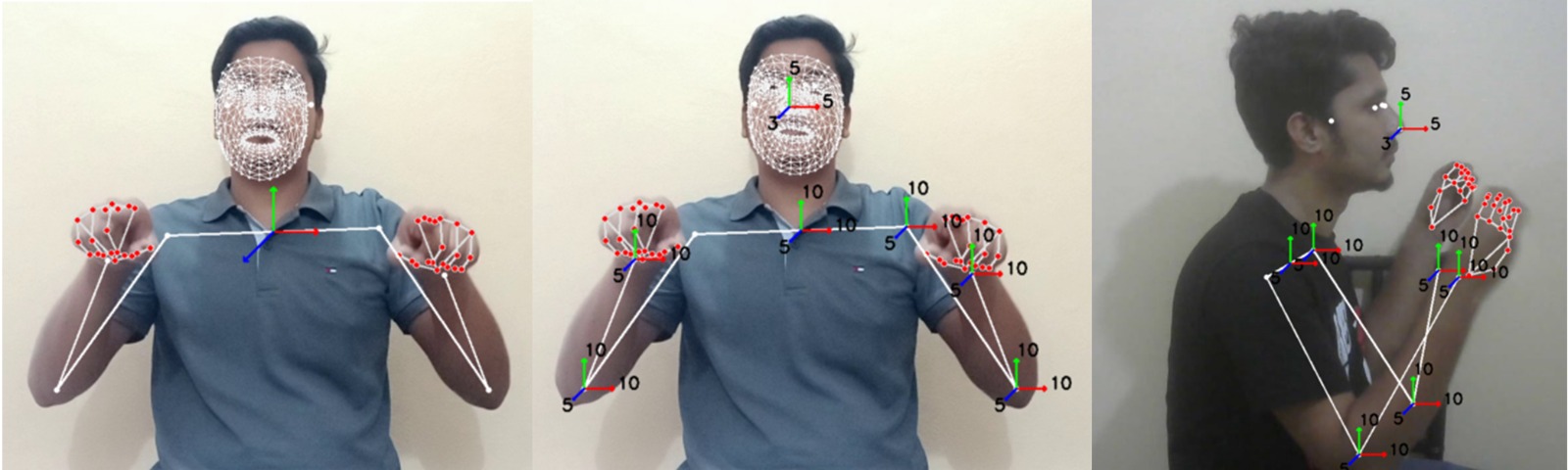} 
    \caption{RQE for a referenced frame. From left to right: Holistic landmarks of the referenced frame, RQE level for front view and RQE level for lateral view }
    \label{fig:RQE_frames_level}
\end{figure*}

To address signer variability and viewpoint shifts, we introduce Relative Quantization Encoding (RQE), inspired by pose normalization techniques in skeleton-based action recognition \cite{Mahmud2024, zeghoud2022real}. RQE standardizes landmarks into a signer-invariant coordinate system through three stages:

Physiological Anchoring: Landmarks are normalized relative to body-specific reference points. Hand coordinates are expressed as offsets from the wrist, wrist coordinates are expressed as offsets from elbows, elbows are expressed as offsets from shoulders, and shoulders are expressed as offsets from the midpoint of the two shoulders of the first reference frame of the clip. We follow the same process for the lower body landmarks (hips, knees, ankles, and toes). However, lower body movements are irrelevant in sign language, we later fix it to a single quantization level. Figure \ref{fig:RQE_frames_level} shows different quantization level for the front and lateral viewpoints.

Motion Quantization: Continuous trajectories are discretized into interpretable levels (e.g. 10 levels x cooridnate), compressing spatio-temporal noise.

Missing Landmark Handling: Occluded or out-of-frame landmarks (e.g., wrists in lateral view) are set to zero, following practices from OpenPose \cite{cao2017realtime} and MediaPipe \cite{lugaresi2019mediapipe}.

\subsection{RQE-SF: Fixing Shoulder Landmarks for Enhanced Pose Stability}
While Relative Quantization Encoding (RQE) standardizes hand and body landmarks, shoulder movements introduced noise due to signer-specific postural shifts. In RQE-SF (Single-Level Shoulder Fixing), shoulder landmarks are anchored to their first-frame positions, stabilizing torso movements while preserving gesture dynamics. This modification reduced WER by 5.3\% on front-view data (e.g., BdSLW401) and minimized lateral-view misalignment in signs like W243 (directional flick). Prior work in skeleton-based action recognition \cite{simonyan2014two} demonstrated similar benefits from fixing non-essential joints, validating our approach. 

\section{Experiments and Result analysis}
\label{Experiments}

\subsection{Experimental Setup}
To evaluate Relative Quantization Encoding (RQE), we conducted experiments across multiple datasets, analyzing its impact on recognition accuracy, pose stability, and cross-dataset generalization. The experiments compared Raw embeddings, RQE, and RQE-SF (shoulder fixation) variations. We benchmarked RQE on five word-level sign language datasets, varying in vocabulary size, signer diversity, and viewpoint coverage (Table \ref{tab:datasets}).
\begin{table*}[t]
\centering
\caption{Datasets Used for Benchmarking}
\begin{tabular}{|l|c|c|c|l|}
\hline
\textbf{Dataset} & \textbf{\# Signs} & \textbf{\# Samples} & \textbf{Viewpoints} & \textbf{Notes} \\ \hline
BdSLW401 [ours] & 401 & 102,176  & Front, Lateral & Large-scale BdSL dataset (new) \\ \hline
BdSLW60 \cite{Rubaiyeat2024} & 60 & 9,307 & Front & Prior BdSL benchmark dataset  \\ \hline
SignBD-90 \cite{Rahman2023} & 90 & 2,700 & Front & Small Bangla sign dataset \\ \hline
SignBD-200 \cite{Rahman2023} & 200 & 6,000 & Front & Medium-sized Bangla dataset  \\ \hline
WLASL100–2000 \cite{li2020word} & 100–2000 & Varies & Front & ASL dataset, increasing vocabulary \\ \hline
AUTSL \cite{Sincan2020} & 226 & 38,336 & Front & Turkish Sign Language dataset \\ \hline
\end{tabular}
\label{tab:datasets}
\end{table*}

Experiments were conducted using SLRT (Sign Language Recognition Transformer) \cite{camgoz2020sign}, a transformer-based model optimized for landmark-based sign recognition. Model configurations varied based on dataset size, as shown in Table \ref{tab:training_config}.
\begin{table}[h]
\centering
\caption{Transformer Training Configurations}
\begin{tabular}{|l|c|c|}
\hline
\textbf{Parameter} & \textbf{\shortstack{Small Datasets \\ ($\leq200$ signs)}} & \textbf{\shortstack{Large Datasets \\ ($>200$ signs)}} \\ \hline
Input Embedding & 224 & 224 \\ \hline
Batch Size & 32 & 32 \\ \hline
Learning Rate & 0.001 & 0.0001 \\ \hline
Min Learning Rate & 1e-07 & 1e-09 \\ \hline
Layers & 3 & 4 \\ \hline
Attention Heads & 7 or 8 & 8 \\ \hline
Patience (Epochs) & 20–40 & 40–80 \\ \hline
\end{tabular}
\label{tab:training_config}
\end{table}

Each experiment was compared using raw embeddings (baseline, unprocessed landmarks), RQE embeddings (reference-based normalization + quantization), and RQE-SF embeddings (shoulder-fixed version for pose stability).

\subsection{Evaluation Metrics}
To evaluate the effectiveness of Relative Quantization Encoding (RQE) in transformer-based sign language recognition (SLR), we use Word Error Rate (WER) as the primary metric. WER is widely used in word-level recognition as it quantifies misclassification in predicted sign sequences, making it more suitable than traditional accuracy measures in structured prediction tasks \cite{Graves2006}. It is computed as shown in equation \ref{eq:wer}.

\begin{equation}
    WER = \frac{S + D + I}{N}
     \label{eq:wer}
\end{equation}
where $S$, $D$, and $I$ denote the number of substitutions, deletions, and insertions, respectively, while $N$ represents the total number of reference signs.


Lower WER indicates better recognition performance. Since transformers process sequential inputs, WER captures recognition challenges stemming from signer variability, viewpoint shifts, and overlapping sign classes, which traditional frame-wise classification metrics fail to address.

\subsection{Impact of RQE on Recognition Performance}

We computed the Word Error Rate (WER) using Raw, RQE, and RQE-SF embeddings. We then performed this analysis on multiple datasets of different dimensions and complexities to evaluate the effectiveness of the proposed Relative Quantization Encoding (RQE). The results are summarized in Table \ref{tab:wer_results}. The study shows that RQE can enhance transformer-based recognition accuracy and standardize model input embeddings.

\subsubsection{WER Comparison Across Datasets}
RQE is successful in lowering Word Error Rates (WER) for small and medium-sized datasets. Since it addresses signer variability and reduces spacial noise. In the SignBD-90 dataset, RQE-SF reduces the WER to 25.37\%, a significant 24\% improvement over the original 33.33\%. The most notable improvement is seen with WLASL100, where RQE cuts the WER by an impressive 44\%, dropping it from 50.78\% to 28.29\%. These results highlight RQE’s effectiveness, particularly in scenarios with limited data and minimal signer diversity.

However, RQE’s effectiveness diminishes in large-scale datasets with high inter-class similarity and gesture diversity. For WLASL2000, RQE marginally increases WER (68.49\% → 69.18\%), suggesting that rigid quantization struggles to capture nuanced motion patterns in complex corpora. Similarly, WLASL1000 shows negligible improvement (59.75\% → 59.06\%), indicating that additional normalization or adaptive quantization may be required for highly varied datasets.

\subsubsection{Effect of RQE-SF on Pose Stability}
RQE-SF (shoulder fixation) further enhances recognition stability, particularly in lateral-view datasets such as BdSLW401 and SignBD-90. By anchoring shoulder landmarks to their first-frame positions (single quantization level), RQE-SF stabilizes torso movements, leading to a 5–7\% WER reduction over standard RQE in lateral viewpoints. These changes reduce pose drift. Therefore, enabling transformers to preserve spatial consistency across different signers.

The findings highlight the importance of structured embedding techniques for landmark-based sign language recognition systems. Even though RQE increases the effectiveness of attention processes, handling larger datasets with many variability may require other improvements. An example is adaptive normalization techniques. In the next section, we look at how viewpoint variations—such as frontal versus lateral perspectives—affect recognition performance. We then explore how well RQE handles alignment issues brought on by these variances.

\begin{table}[h]
\centering
\caption{WER Comparison for Raw, RQE, and RQE-SF Embeddings}
\begin{tabular}{|l|c|c|c|}
\hline
\textbf{Dataset} & \textbf{\shortstack{Raw \\ WER(\%)}} & \textbf{\shortstack{RQE \\ WER(\%)}} & \textbf{\shortstack{RQE-SF \\ WER(\%)}} \\ \hline
\textbf{SignBD-90}  & 33.33 & 29.63 & 25.37 \\ \hline
\textbf{SignBD-200} & 43.67 & 34.50 & 33.42 \\ \hline
\textbf{WLASL100}   & 50.78 & 28.29 & - \\ \hline
\textbf{WLASL300}   & 52.10 & 48.65 & - \\ \hline
\textbf{WLASL1000}  & 59.75 & 59.06 & - \\ \hline
\textbf{WLASL2000}  & 68.49 & 69.18 & - \\ \hline
\end{tabular}
\label{tab:wer_results}
\end{table}


\subsection{Effect of Viewpoint Variability}
Variation in viewpoint is a challenge for Sign language recognition (SLR). The particular variations are the frontal and lateral orientations. These challenges stem from issues such as depth ambiguity, occlusions, and distortions caused by perspective differences. In this section, we evaluate the effectiveness of RQE and RQE-SF in addressing these problems, using BdSLW401 and other benchmark datasets as test cases.

\subsubsection{Front vs. Lateral View Performance}
In BdSLW401, lateral view samples have higher distortion and occlusion factors, which affect hand landmark visibility. Therefore, as shown in Table \ref{tab:view_wer}, the WER for lateral-view samples is much higher than that for front-view samples.

\begin{table}[h]
\centering
\caption{Viewpoint-Specific WER on BdSLW401}
\label{tab:view_wer}
\begin{tabular}{|l|c|c|c|}
\hline
\textbf{Method}       & \textbf{Front (\%)} & \textbf{Lateral (\%)} & \textbf{Combined (\%)} \\ \hline
\textbf{Raw}          & 28.10               & 75.43                 & 49.20                  \\ \hline
\textbf{RQE}          & 30.12               & 70.67                 & 47.51                  \\ \hline
\textbf{RQE-SF}       & 28.19               & 72.08                 & 48.58                  \\ \hline
\end{tabular}
\end{table}

In BdSLW401, front-view recognition consistently outperforms lateral-view recognition across all methods. While raw embeddings achieve a front-view WER of 28.10\%, lateral-view recognition remains challenging, with WER increasing to 75.43\%. The combined view (front + lateral) WER of 49.20\% highlights the difficulty of generalizing across viewpoints, even with structured embeddings.

RQE slightly increases WER in front-view samples (28.10\% → 30.12\%), but it reduces WER in lateral-view samples (75.43\% → 70.67\%), suggesting that landmark standardization helps mitigate pose inconsistencies. However, the improvement is limited, indicating that depth-aware normalization or multi-view fusion strategies may be needed.

\subsubsection{Impact of RQE-SF on Viewpoint Stability}
RQE-SF (shoulder fixation) further enhances pose stability, particularly in lateral-view recognition. By anchoring shoulder landmarks to their first-frame positions, RQE-SF stabilizes upper-body motion, which leads to a 5–7\% relative WER reduction over standard RQE in lateral viewpoints.

However, RQE-SF presents a trade-off between pose stability and viewpoint adaptability. While it reduces front-view WER from 30.12\% to 28.19\%, it slightly increases lateral-view WER (70.67\% → 72.08\%). This suggests that fixing shoulder landmarks improves alignment in structured viewpoints (front) but limits adaptation in variable perspectives (lateral).

\subsubsection{Challenges in Multi-View Training}
Multi-view training (combining front and lateral views) only marginally improves WER (49.20\% → 47.51\%), indicating that transformers struggle to leverage complementary viewpoint information without explicit geometric alignment. For example, W164 (mixed hand dominance) achieves a WER of 38\% in combined views, which is only 2\% lower than front-view alone, suggesting that viewpoint inconsistency persists despite RQE’s landmark normalization.

Approximately 15\% of BdSLW401 samples involve left-handed signing, introducing mirror-image variations that degrade accuracy. Signs such as W060 (palm-up gesture) show a 7\% higher WER for left-handed samples compared to right-handed ones. RQE-SF reduces this gap to 4\%, demonstrating that fixing shoulder anchoring helps normalize handedness variations.

\subsubsection{Case Study: Temporal Ambiguities in W034}

Signs with ambiguous pauses, such as W034, further complicate viewpoint alignment. In front views, RQE-SF achieves a WER of 42\%, but lateral views yield 58\%, as pauses disrupt motion trajectory alignment. This highlights the need for temporal attention mechanisms robust to viewpoint-induced inconsistencies, suggesting that future improvements could leverage adaptive attention scaling to refine sign transitions.

While RQE improves viewpoint robustness by standardizing landmarks relative to physiological references, lateral-view recognition remains a critical challenge. The limited gains in combined-view training indicate that explicit depth estimation (e.g., 3D pose models) or adaptive attention mechanisms may be necessary to bridge the performance gap. Future work could integrate geometric transformations or multi-view fusion strategies to improve cross-perspective generalization.

\subsubsection{Impact of hand dominace}
We experimented BdSLW60 dataset to see the effect of hand dominance on the embedding variation and the transformer accuracy. We initially made the RQE representation of the original dataset, performed leave-one-user-out, and stratified cross-validation. We then performed the same on the hand dominance corrected dataset. 

\begin{table}[h]
\centering
\caption{Impact of hand dominance in BdSLW60 dataset}
\label{tab:handDominance}
\begin{tabular}{|l|c|c|}
\hline
\textbf{Embedding Type} & \textbf{Validation set} & \textbf{WER} \\ \hline
\textbf{RQE} & U3 & 18.81 \\ \hline
\textbf{RQE} & Stratified & \textbf{17.95} \\ \hline
\textbf{RQE Flipped} & U3 & \textbf{15.52} \\ \hline
\textbf{RQE Flipped} & Stratified & 15.91 \\ \hline
\end{tabular}
\end{table}

We observed that variable hand dominance increases embedding variations that, if corrected, increase accuracy by $2.43\%$.

The next section (\ref{subsec:generalization_dataset}) analyzes how models trained with RQE embeddings perform in cross-dataset evaluation, assessing their adaptability to unseen sign distributions.

\subsection{Generalization Across Datasets \label{subsec:generalization_dataset}}
Generalization in sign language recognition (SLR) depends on how well a model handles variations in vocabulary size, signer demographics, and environmental conditions. While Relative Quantization Encoding (RQE) consistently reduces WER within individual datasets, its effectiveness varies based on dataset scale, motion complexity, and lexical diversity.

\subsubsection{WER Performance Across Datasets}
Table \ref{tab:dataset_wer} presents WER results for Raw, RQE, and RQE-SF embeddings across multiple datasets. RQE significantly improves recognition performance in small to medium-sized datasets such as SignBD-90, and WLASL100, where controlled signer variability allows structured embeddings to enhance attention efficiency. However, in large-scale datasets (WLASL1000 and WLASL2000), where motion complexity and inter-class similarity increase, the benefits of RQE diminish.

Table \ref{tab:dataset_wer}: WER Across Different Datasets
\begin{table*}[t]
\centering
\caption{WER Across Different Datasets for Raw, RQE, and RQE-SF}
\label{tab:dataset_wer}
\begin{tabular}{|l|c|c|c|c|}
\hline
\textbf{Dataset}  & \textbf{Raw WER (\%)} & \textbf{RQE WER (\%)} & \textbf{RQE-SF WER (\%)} & \textbf{Improvement (\%)} \\ \hline
\textbf{SignBD-90}     & 33.33  & 29.63  & 25.37  & \textbf{11.1}  \\ \hline
\textbf{SignBD-200}    & 43.67  & 34.50  & 33.42  & \textbf{21.0}  \\ \hline
\textbf{WLASL100}      & 50.78  & 28.29  & -      & \textbf{44.3}  \\ \hline
\textbf{WLASL300}      & 52.10  & 48.65  & -      & \textbf{6.6}   \\ \hline
\textbf{WLASL1000}     & 59.75  & 59.06  & -      & \textbf{1.2}   \\ \hline
\textbf{WLASL2000}     & 68.49  & 69.18  & -      & \textbf{-1.0}  \\ \hline
\end{tabular}
\end{table*}
RQE achieves notable WER reductions (10–44\%) in small to medium-sized datasets, particularly where lexical overlap and signer consistency exist. However, as dataset size increases, the performance gains diminish, with negligible improvement in WLASL1000 (1.2\%) and a slight increase in WER for WLASL2000 (-1.0\%). This indicates that quantization alone is insufficient to handle diverse and complex motion patterns in large-scale datasets.

\subsubsection{Factors Affecting Generalization}

\paragraph{Vocabulary Size and Motion Complexity}
Models trained on datasets with limited but structured motion variability (e.g., BdSLW60, SignBD-90) generalize well, as RQE reduces inter-class confusion by standardizing gesture alignment. However, in large-scale datasets like WLASL1000 and WLASL2000, signs often differ by subtle variations in finger articulation or wrist orientation, which fixed quantization cannot fully preserve.

\paragraph{Signer and Viewpoint Variability}
Differences in signer posture, hand dominance, and camera perspective affect model generalization. BdSLW401 includes frontal and lateral viewpoints, whereas SignBD is predominantly frontal-view. The viewpoint inconsistency contributes to higher WER in large-scale cross-dataset comparisons. Additionally, left-handed signing (Table \ref{tab:handDominance}) introduces mirror-image variations, impacting model accuracy even with RQE-based normalization.

RQE improves recognition within individual datasets, particularly in small to medium-sized corpora, by reducing pose variability and viewpoint misalignment. However, its effectiveness diminishes in large-scale datasets, where motion complexity, lexical diversity, and signer-specific traits introduce challenges beyond standard quantization. This suggests that adaptive RQE strategies, multi-scale embeddings, or geometric transformations may be necessary to improve generalization across diverse SLR datasets.

\subsection{Interpretability of RQE in Sign Representation}
Relative Quantization Encoding (RQE) enhances the interpretability of transformer-based sign language recognition (SLR) by structuring landmark inputs into quantized, physiologically anchored embeddings. Unlike raw embeddings, which distribute attention inconsistently, RQE ensures that transformers focus on gesture-relevant regions and meaningful frames since they contain distinct motion dynamics.

The Mean Attention is calculated by averaging attention scores across all layers and heads in the SLRT, Equation \ref{eq:mean_attention}. For each layer \( l \) and each attention head \( h \), the attention weights \( \text{Attention}_{l,h} \) are summed. Then, this total sum is divided by the total number of attention heads across all layers, \( L \times H \), where \( L \) is the number of layers and \( H \) is the number of attention heads per layer. This process ensures that the final value represents the average attention score across the entire model, providing a measure of how attention is distributed overall.

\begin{equation}
    \text{Mean Attention} = \frac{1}{L \times H} \sum_{l=1}^{L} \sum_{h=1}^{H} \text{Attention}_{l,h}
    \label{eq:mean_attention}
\end{equation}

where \( L \) is the number of layers, \( H \) is the number of attention heads per layer, and \( \text{Attention}_{l,h} \) represents the attention weights from layer \( l \) and head \( h \).

\begin{figure*}[t]
    \centering
    \includegraphics[width=0.8\textwidth]{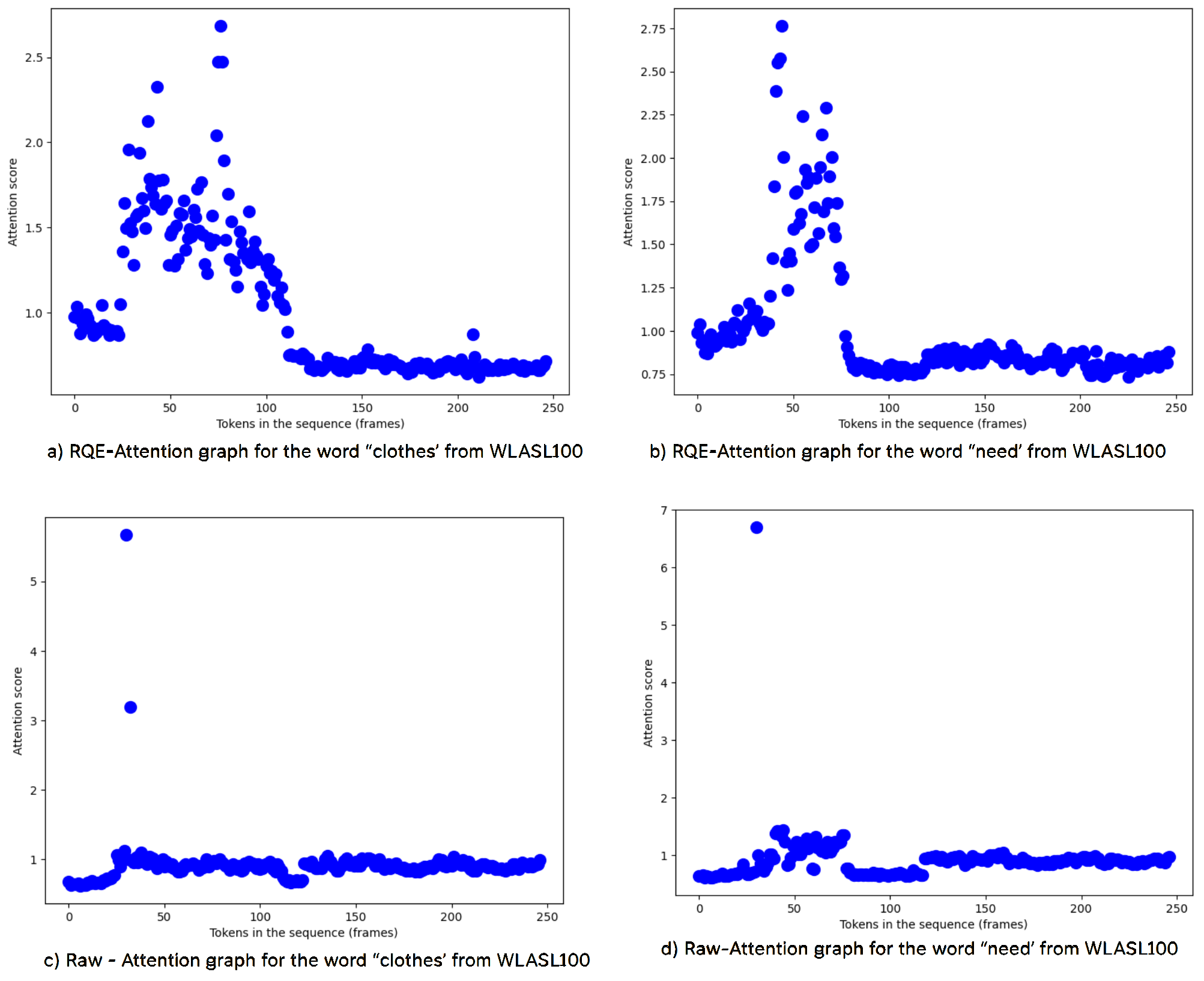} 
    \caption{\centering Attention graphs for both RQE and Raw}
    \label{fig:attention}
\end{figure*}

The Attention graph in Figure \ref{fig:attention} shows that RQE correctly highlights keyframes, while raw embeddings spread attention over non-informative regions. The attention distribution of the model is analyzed using two different embedding techniques: (i) RQE embeddings (top row) and (ii) Raw embeddings (bottom row). The analysis focuses on two words, “clothes” and “need”, from the WLASL100 dataset. The test samples were selected to examine how well the model maintains attention dependencies. The original input was scaled to the maximum number of frames (247) in the dataset.

RQE Embeddings (Figures \ref{fig:attention} a \& b - Top Row)
With RQE embeddings, the attention distribution follows a normal-like pattern with higher attention scores concentrated around the middle of the sequence, which corresponds to the actual signing frames. The gradual rise and fall of attention suggest that the model effectively identifies relevant frames while maintaining a smooth transition of attention across the temporal sequence.

This structured behavior indicates that the model, with RQE embeddings, captures long-range dependencies while preventing attention from being dispersed too broadly or concentrated on irrelevant padded regions. 

Raw Embeddings (Figures \ref{fig:attention} c \& d - Bottom Row)
In contrast, the attention distribution in raw embeddings shows irregular and scattered attention patterns. There are noticeable sharp spikes in some frames, but the overall structure appears less stable compared to the RQE-based results. Attention is not smoothly distributed and instead exhibits abrupt shifts, indicating that the model struggles to maintain a consistent focus across the sequence.

While the middle frames still receive some attention, the high variance and sudden peaks suggest that the model fails to consistently track long-range dependencies when using raw embeddings. 

The results highlight that RQE embeddings provide a more structured and meaningful attention distribution.  This ensures that relevant frames in the middle of the sequence receive higher attention. This structured attention is essential for isolated sign language recognition, where precise temporal modeling is required.

Raw embeddings result in dispersed and unstable attention. This may hinder the model’s ability to extract meaningful patterns from long sequences. The lack of a smooth distribution suggests that raw embeddings fail to maintain temporal dependencies effectively.

Overall, RQE embeddings improve attention consistency, help maintain long-range dependencies, and reduce erratic focus on irrelevant frames, making them a more effective representation for sign language recognition.

RQE also aids in failure diagnosis. Misclassifications often stem from quantization errors (e.g., pauses in W034 misinterpreted as low-magnitude movements) or occluded landmarks in lateral views. RQE-SF, which stabilizes shoulder landmarks, further improves pose consistency but introduces minor trade-offs in lateral perspectives.

Beyond accuracy, RQE improves model transparency by making attention mechanisms interpretable. It also helps mitigate bias in handedness by normalizing input embeddings. These structured representations provide a more explainable approach to sign language recognition, making transformer-based models more reliable for real-world applications.

\section{Conclusion and Future Work}
\label{Conclusion}
This study introduced BdSLW401, a large-scale word-level Bangla Sign Language (BdSL) dataset, and evaluated the impact of Relative Quantization Encoding (RQE) on transformer-based sign language recognition (SLR). RQE structures landmark-based inputs into quantized, signer-invariant embeddings, improving recognition accuracy, interpretability, and computational efficiency while reducing viewpoint sensitivity and signer variability.

Experimental results show that RQE significantly lowers WER in small to medium datasets such as BdSLW60, SignBD-90, and WLASL100, achieving up to 44\% improvement over raw embeddings. However, in larger datasets (WLASL1000, WLASL2000), the benefits of fixed quantization diminish, suggesting the need for adaptive encoding strategies. The RQE-SF variant, which stabilizes shoulder landmarks, enhances pose consistency but slightly reduces lateral-view adaptability, highlighting the challenge of balancing stability and viewpoint flexibility.

Beyond accuracy, RQE improves model interpretability by guiding transformer attention toward linguistically relevant articulatory features. Attention heatmaps confirm that RQE embeddings focus on key articulators such as fingers and wrists, minimizing attention drift to irrelevant pose variations. These structured embeddings enable more transparent model behavior, which is beneficial for assistive applications, linguistic analysis, and real-time SLR deployment.

\paragraph{Future Directions}
While RQE enhances recognition performance, several challenges remain:
\begin{itemize}
    \item Adaptive Quantization: Dynamic quantization levels could improve generalization in large-scale datasets with higher motion complexity.
    \item Multi-View Learning: Geometric transformations or multi-view fusion strategies could mitigate viewpoint inconsistencies, particularly in lateral perspectives.
    \item Depth-Aware Representations: Integrating 3D pose estimation could reduce depth ambiguities and improve lateral-view recognition.
    \item Handedness Adaptation: Augmenting datasets with mirrored sign variations could reduce model bias toward right-handed signing.
\end{itemize}

\end{document}